\documentclass[sigconf]{acmart}
\usepackage{bm}
\usepackage{mathtools}
\usepackage{adjustbox}
\usepackage{multirow}
\usepackage{enumitem}
\usepackage{hyperref}

\AtBeginDocument{%
  \providecommand\BibTeX{{%
    \normalfont B\kern-0.5em{\scshape i\kern-0.25em b}\kern-0.8em\TeX}}}

\begin{document}

\title{Exploring Multi-view Pixel Contrast for General and Robust Image Forgery Localization}

\author{Zijie Lou$^{1,2}$, Gang Cao$^{1,2,*}$, Kun Guo$^{1,2}$, Haochen Zhu$^{1,2}$, Lifang Yu$^{3}$}
\thanks{*Corresponding author.}
\affiliation{%
	\institution{$^{1}$State Key Laboratory of Media Convergence and Communication, Communication University of China}
    \institution{$^{2}$School of Computer and Cyber Sciences, Communication University of China}
    \institution{$^{3}$Department of Information Engineering, Beijing Institute of Graphic Communication}
	\city{Beijing}
	\country{China}
}
\email{gangcao@cuc.edu.cn}

\renewcommand{\shortauthors}{Zijie Lou and Gang Cao, et al.}
\begin{abstract}
Image forgery localization, which aims to segment tampered regions in an image, is a fundamental yet challenging digital forensic task. While some deep learning-based forensic methods have achieved impressive results, they directly learn pixel-to-label mappings without fully exploiting the relationship between pixels in the feature space. To address such deficiency, we propose a Multi-view Pixel-wise Contrastive algorithm (MPC) for image forgery localization. Specifically, we first pre-train the backbone network with the supervised contrastive loss to model pixel relationships from the perspectives of within-image, cross-scale and cross-modality. That is aimed at increasing intra-class compactness and inter-class separability. Then the localization head is fine-tuned using the cross-entropy loss, resulting in a better pixel localizer. The MPC is trained on three different scale training datasets to make a comprehensive and fair comparison with existing image forgery localization algorithms. Extensive experiments on the small, medium and large scale training datasets show that the proposed MPC achieves higher generalization performance and robustness against post-processing than the state-of-the-arts. Code will be available at \href{https://github.com/multimediaFor/MPC}{https://github.com/multimediaFor/MPC}.
\end{abstract}

\maketitle

\section{Introduction}
Digital images can be easily manipulated owing to the rapid development of image processing techniques and tools. Malicious image manipulation, such as fake news, academic fraud and criminal activities, can have a great negative impact on society. Moreover, the tampered regions are often indistinguishable to the naked eye due to the imperceptible tampering artifacts. Therefore, it is crucial to develop reliable image manipulation forensic techniques.

Early methods mainly focus on using handcrafted features to distinguish pristine and tampered regions, such as color filter array \cite{ferrara2012image} and local noise features \cite{mahdian2009using}. However, the handcrafted features cannot generalize well on unseen manipulation types due to the reliance on prior statistical models. In recent years, deep learning (DL)-based methods \cite{hu2020span, liu2022pscc, salloum2018image, bi2019rru, li2019localization, bappy2019hybrid, kwon2022learning, liu2023evp, dong2022mvss, kong2023pixel, gao2022tbnet, rao2022towards, wu2022robust, wang2022objectformer, zhuo2022self, hao2021transforensics, guillaro2023trufor, zhou2018learning, zhou2020generate, shi2023transformer, xu2023up, li2022image, liu2023tbformer, lin2023image, guo2023hierarchical, li2023edge, zeng2023towards, zhuang2021image, zhang2024catmullrom, zhu2024learning, zeng2024mgqformer} have achieved impressive performance on image forgery localization. With abundant training samples and powerful feature representation capability, such DL-based methods significantly outperform the traditional ones in terms of localization performance. Typical image forgery localization models consist of a deep feature extractor followed by a pixel-wise softmax/sigmoid classifier, and are trained using a pixel-wise cross-entropy (CE) loss. To effectively distinguish between tampered and real regions, some forensic methods adopt specialized network designs, \textit{e.g.}, multi-layer feature fusion \cite{dong2022mvss, gao2022tbnet, guillaro2023trufor, kong2023pixel, li2022image, liu2022pscc, liu2023tbformer, hu2020span, shi2023transformer} and attention mechanism \cite{dong2022mvss, guillaro2023trufor, guo2023hierarchical, hao2021transforensics, hu2020span, kong2023pixel, li2023edge, liu2022pscc, liu2023tbformer, wang2022objectformer, zhuo2022self}. Basically, these forgery localization models utilize deep networks to project image pixels into a highly non-linear feature space. However, they usually learn pixel-to-label mappings in the label space directly, but ignore the relationships between pixels in the feature space. Ideally, an effective forgery localization feature space should not only 1) consider the categorization capability of individual pixel embeddings, but also 2) possess a well-organized structure to handle both intra-class compactness and inter-class separability.

To address this issue, we propose a multi-view pixel-wise contrastive algorithm for more effective image forgery localization. Specifically, we first pre-train the backbone network using supervised contrastive loss from three perspectives: within-image, cross-scale, and cross-modality to shape the pixel feature space. And then the localization head is fine-tuned using the CE loss to address inter-class discrimination issues. This approach increases both intra-class compactness and inter-class separability, as the contrastive loss encourages the features of pixels from the same class (pristine and pristine, tampered and tampered) to be close to each other and the features of pixels from different classes (pristine and tampered) to be far away. This naturally lead to more accurate localization predictions in the fine-tuning stage.

Some recent works such as \cite{zeng2023towards, niloy2023cfl} also used contrastive learning for the task of image forgery localization. However, these works only focus on within-image pixel contrast, train with contrastive loss and CE loss at the same time, and validate their schemes on limited test datasets. In comparison with these works, we utilize contrastive loss to explore the structural information of labeled pixel embeddings from three perspectives: within-image, cross-scale, and cross-modality. In addition, we adopt a two-stage training strategy and demonstrate the effectiveness of proposed method on numerous test datasets. Overall, our contributions are as follows: 

\begin{itemize}
\item{We propose a multi-view pixel-wise contrastive algorithm (MPC) for image forgery localization. The two-stage training strategy enables the proposed method to model not only pixel relationships in the label space but also in the feature space.}
\item{Contrastive losses are used to shape pixel feature space from three perspectives: within-image, cross-scale, and cross-modality. A well structured pixel feature space is obtained by the multi-view pixel contrast, thus improving the localization performance.} 
\item{We conduct a comprehensive and fair comparison with existing image forgery localization methods. Extensive experiments show that proposed MPC achieves significant performance compared to state-of-the-art methods.}
\end{itemize}

\section{Related Works}
\subsection{Image Forgery Localization}
Traditional methods \cite{ferrara2012image, mahdian2009using, fan2015image, carvalho2015illuminant} for image forgery localization mainly rely on handcrafted features to capture the statistical anomalies incurred by tampering operations. Early work \cite{ferrara2012image} utilized color filter array to detect the inconsistency of the tampered regions in an image. Moreover, the local noise features introduced by the sensors \cite{mahdian2009using} and the inconsistencies of the illuminant color or lighting \cite{fan2015image, carvalho2015illuminant} are also clues for image splicing localization. However, such handcrafted features are defined for specific types of image forgeries, which are difficult to generalize well on unseen manipulation types in practice. 

Recently, extensive DL-based architectures are adopted to extract the forensic features adaptively, leading to better generalization ability in face of various manipulations. The employed backbones include convolutional neural network (CNN) \cite{zhou2018learning, li2023edge, dong2022mvss, guo2023hierarchical, wu2022robust, kwon2022learning, liu2022pscc, hu2020span, zhuo2022self, li2022image, zeng2023towards, xu2023up}, long short-term memory (LSTM) network \cite{bappy2019hybrid}, fully convolutional network (FCN) \cite{wu2019mantra, salloum2018image, zhuang2021image, li2019localization} and Transformer \cite{wang2022objectformer, guillaro2023trufor, hao2021transforensics, liu2023tbformer, shi2023transformer, liu2023evp, zeng2024mgqformer}. Siamese network \cite{huh2018fighting, cozzolino2019noiseprint} is also used to identify the tampered regions by exploring the patch consistency. Such methods rely on different camera attributes, thus the images with meta-information about camera are required. In contrast, we use the tampered images with only pixel information. In this work, we do not design more complex network architectures. Instead, we focus on shaping a more structured feature space to achieve better localization performance.

It is noteworthy that existing methods employ differing scales of training datasets, making fair comparisons difficult to achieve. The common training dataset scales can be categorized into three types: small-scale (approximately 5000 samples), medium-scale (about 100,000 samples), and large-scale (exceeding 800,000 samples). For example, MVSS-Net++ \cite{dong2022mvss}, PSCC-Net \cite{liu2022pscc}, and TruFor \cite{guillaro2023trufor} utilized approximately 5000, 100000, and 800000 tampered images to train their networks, respectively. Therefore, unless retraining with datasets of identical or equivalent scale, simultaneous comparisons should be avoided. To make a comprehensive and fair comparison with existing image forgery localization algorithms, we train our network with identical (\textbf{Experiment 1} and \textbf{Experiment 3}) or equivalent scale datasets (\textbf{Experiment 2}) in such three experimental setups, as shown in Table \ref{dataset}. Additionally, due to the lack of publicly available code for many methods \cite{kong2023pixel, xu2023up, shi2023transformer, zhuo2022self, li2022image, li2023edge, lin2023image, zhang2024catmullrom, zeng2024mgqformer, zhu2024learning}, we directly cite the results from their respective literatures after aligning the training datasets.

\begin{table}[!t]
\caption{Training / Testing split and characteristic of public benchmark datasets. CASIA contains two versions that CASIAv1 and CASIAv2 contain 920 and 5123 manipulated images, respectively. Abbreviations: S – Splicing; C – Copy-move; R – Removal; L – Locally AI-generation; D – DeepFake; PT – Pre-Training; FT – Fine-Tuning.}
\tabcolsep=2.5 pt
\begin{adjustbox}{width=\linewidth}
\begin{tabular}{r|c|cc|cc|cc}
\toprule[1.5pt]
\multirow{2}{*}{Dataset} & \multirow{2}{*}{Type}     & \multicolumn{2}{c|}{\textbf{Experiment 1}} & \multicolumn{2}{c|}{\textbf{Experiment 2}} & \multicolumn{2}{c}{\textbf{Experiment 3}} \\ \cmidrule{3-8}
                                                    &              & Training     & Testing          & Training     & Testing         & Training     & Testing              \\         
\midrule                    
CASIA \cite{dong2013casia}                      &S, C           &5123           &920             &5123 (FT)          &920          &5123         &920     \\
NIST \cite{guan2019mfc}                       &S, C, R        &-              &564             &404 (FT)           &160          &-            &564         \\
Columbia \cite{hsu2006columbia}                   &S              &-              &160             &-                  &-          &-            &160            \\
IFC \cite{IFC}                        &-              &-              &450             &-                  &-            &-            &-             \\
IMD \cite{novozamsky2020imd2020}                        &S, C, R        &-              &2010            &-                  &2010         &2010         &-              \\
Coverage \cite{wen2016coverage}                   &C              &-              &100             &75 (FT)            &25           &-            &100               \\ 
DSO \cite{carvalho2015illuminant}                        &S              &-              &100             &-                  &-            &-            &100               \\
DEF-test \cite{mahfoudi2019defacto}                   &S, C, R        &-              &6000            &-                  &-            &-            &-              \\
Wild \cite{huh2018fighting}                       &S              &-              &201             &-                  &-          &-            &201            \\ 
Korus \cite{korus2016evaluation}                      &-              &-              &220             &-                  &-            &-            &220            \\
MISD \cite{kadam2021multiple}                       &S              &-              &-               &-                  &-            &-            &227            \\
CoCoGlide \cite{guillaro2023trufor}          &L              &-              &-               &-                  &-            &-            &512            \\
FF++  \cite{rossler2019faceforensics++}        &D              &-              &-               &-                  &-            &-            &1000            \\
SP-COCO \cite{kwon2022learning}                    &S              &-              &-               &20\textit{k} (PT)  &-            &200\textit{k}&-            \\
CM-COCO \cite{kwon2022learning}                    &C              &-              &-               &20\textit{k} (PT)  &-            &200\textit{k}&-    \\
CM-RAISE \cite{kwon2022learning}                   &C              &-              &-               &20\textit{k} (PT)  &-            &200\textit{k}&-   \\
CM-C-RAISE \cite{kwon2022learning}                 &C              &-              &-               &-                  &-            &200\textit{k}&-            \\
\bottomrule[1.5pt]
\end{tabular}
\end{adjustbox}
\label{dataset}
\end{table}

\subsection{Contrastive Learning}
Recently, contrastive learning has made important progress in self-supervised learning \cite{he2020momentum, chen2020simple}, which acquires effective representations without supervision. The discriminative representation learning is achieved by contrasting similar (positive) data pairs against dissimilar (negative) pairs. Specifically, augmented versions of an instance are used to form the positive pairs, while negative pairs are usually yielded by random sampling. Moreover, supervised contrastive loss \cite{khosla2020supervised} is proposed to improve the accuracy of image classification.

\begin{figure*}[!t]
\centering
\includegraphics[width=\textwidth]{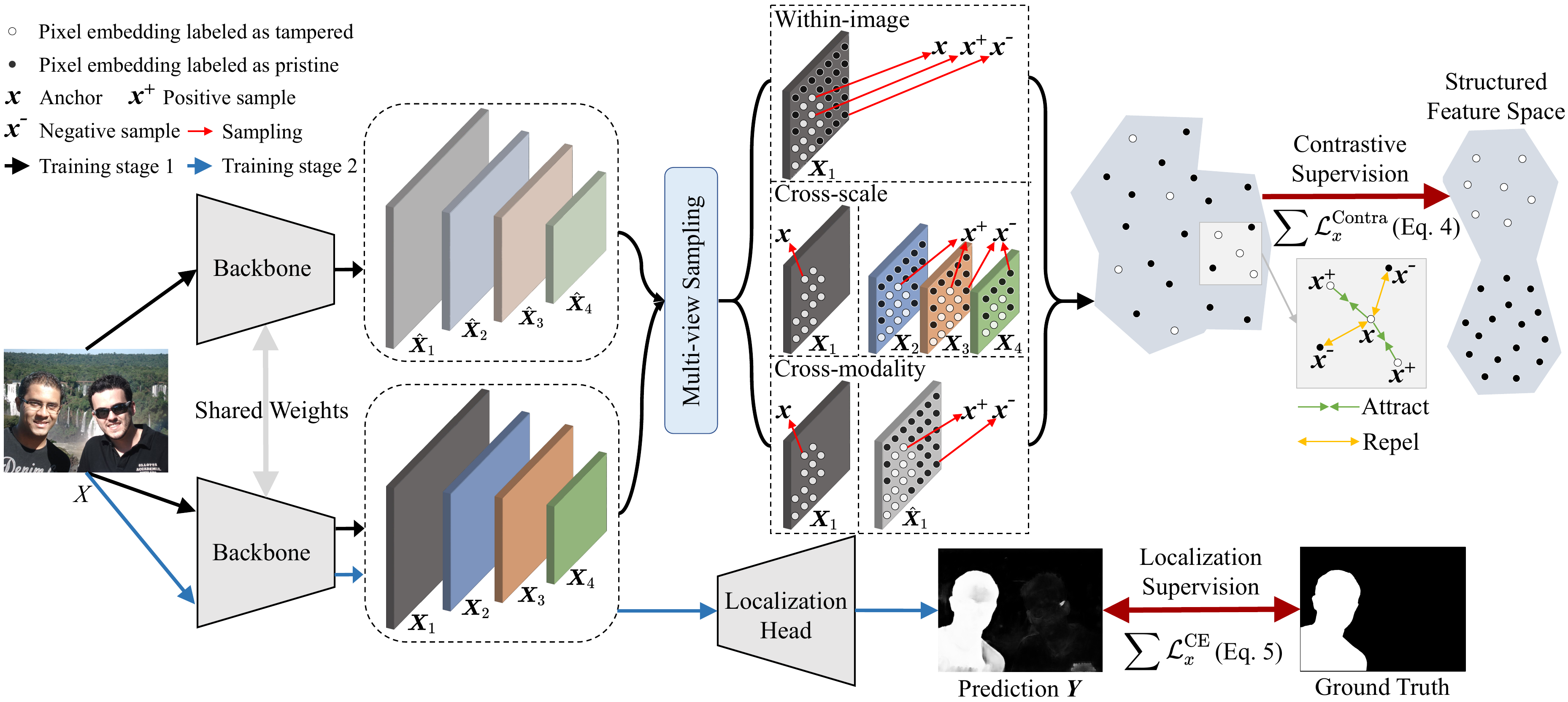}
\caption{Detailed illustration of proposed image forgery localization network MPC.}
\label{MPC}
\end{figure*}

Motivated by the fact that contrastive learning can capture the rich spatial relationships of local features, there are increasingly more researches applying contrastive learning to the field of forensics. For example, \cite{sun2022dual} uses contrastive loss to supplement CE loss for face forgery detection. Noticing the limitations of the widely-used CE loss, some recent works also involve contrastive loss to assist the network training for image forgery localization \cite{niloy2023cfl, zeng2023towards}. However, different from such works only focus on within-image pixel contrast, we utilize contrastive loss to shape the pixel feature space from three perspectives: within-image, cross-scale, and cross-modality. Further, these works train with contrastive loss and CE loss at the same time while we adopt a two-stage training strategy. Also, \cite{niloy2023cfl, zeng2023towards} do not demonstrate the effectiveness of contrastive learning on numerous test datasets. In this work, we substantiate the significant potential of contrastive learning in the task of image tampering localization through extensive experimentation.

\section{Method}
\subsection{Network Architecture} Our algorithm has two major components including the backbone network and localization head, as shown in Fig. \ref{MPC}.

HRFormer \cite{yuan2021hrformer} is employed as the backbone network $f_{BAC}$, which maps an input image $X$ into dense embeddings $\{\bm{X}_i, \hat{\bm{X}}_i\} = f_{BAC}(X)$. We adopt HRFormer in a forensic task for the first time because it maintains high-resolution representations through the entire process. Such high-resolution representations enable the fine-grained forensic clues preserved, which are crucial for accurate forgery localization. In addition, our cross-scale and cross-modality contrastive losses are dependent on the architectural design of HRFormer. The localization head $f_{LOC}$ is implemented by $1 \times 1$ convolutional layers with ReLU, which projects concatenated $\{\bm{X}_i\}$ into a score map $\bm{Y} = f_{LOC}(Concat(\bm{X}_1, \bm{X}_2, \bm{X}_3, \bm{X}_4))$.

It is worth noting that we employ a two-stage training strategy. Specifically, the backbone network is trained using contrastive loss (Eq. \ref{total_loss}) initially, after which the weights of the backbone network are frozen. Subsequently, we fine-tune the localization head using CE loss (Eq. \ref{CE}).

\subsection{Forensics-towards Multi-view Pixel Contrast}
In this work, we develop a multi-view pixel-wise contrastive learning algorithm to enhance the representation learning for forged pixels detection. We cast the image forgery localization task as a pixel-wise binary classification problem, \textit{i.e.}, each pixel $x$ of an image $X$ must be classified into a class $\hat{y} \in \{0, 1\}$, where "0" indicates "pristine" and "1" indicates "tampered". Specifically, given a forged image $X \in \mathbb{R}^{H\times W\times 3}$, the backbone network produces a series of multi-scale features $\{\bm{X}_1, \bm{X}_2, \bm{X}_3, \bm{X}_4\}$ and $\{\hat{\bm{X}_1}, \hat{\bm{X}_2}, \hat{\bm{X}_3}, \hat{\bm{X}_4}\}$. Due to the presence of \textit{dropout} in the backbone, $\bm{X}_i$ and $\hat{\bm{X}}_i$ are different. Thus the pixel embedding $\bm{x} \in \mathbb{R}^{C}$ of $x$ can be derived.

\noindent 
\textbf{Within-image loss.} 
We first explore the structural relationship between pixels within an image. For a pixel embedding $\bm{x} \in \bm{X}_1$ with groundtruth label "1", the positive samples $\bm{x}^+$ are the other pixel embeddings with label "1" in $\bm{X}_1$, while the negative samples $\bm{x}^-$ are the ones with label "0" in $\bm{X}_1$. The within-image contrastive loss is defined as

\begin{align}
\mathcal{L}^{\text{1}}_x= -_{\!}\log\frac{\frac{1}{|\mathcal{P}_x|} \sum_{\bm{x}^+ \in \mathcal{P}_x \in \bm{X}_1} \exp(\bm{x} \cdot \bm{x}^{+}/\tau)}{\sum\nolimits_{\bm{x}^{-}\in\mathcal{N}_x \in \bm{X}_1} \exp(\bm{x} \cdot \bm{x}^{-}/\tau)} \tag{1}
\label{within-image}
\end{align}
\noindent
where $\mathcal{P}_x$ and $\mathcal{N}_x$ denote the positive and negative pixel embedding collections randomly sampled from $\bm{X}_1$, respectively. ‘·’ denotes the inner product, and $\tau$>0 is a temperature hyper-parameter. Such within-image contrastive loss is aimed at learning discriminative feature representation, which benefits to distinguish the pristine and forged pixels within an image. By pulling the same class of pixel embeddings close and pushing different class of pixel embeddings apart, the intra-class compactness and inter-class separability can be improved \cite{niloy2023cfl, wang2021exploring, zhao2021contrastive}.

\noindent 
\textbf{Cross-scale loss.}  
Since tampered regions typically enjoy various sizes, it is important to fuse the features across different scales for enhancing the robustness of localization algorithms against scale variation. Different from \cite{guillaro2023trufor, kong2023pixel} that combine low-level features into the high-level features directly, our cross-scale contrastive loss is computed between different scale features. Specifically, for a pixel embedding $\bm{x} \in \bm{X}_1$, the positive and negative pixel embedding collections $\mathcal{P}_x$, $\mathcal{N}_x$ are sampled from $\bm{X}_2$, $\bm{X}_3$ and $\bm{X}_4$. That is

\begin{align}
\mathcal{L}^{\text{2}}_x= -_{\!}\log\frac{\frac{1}{|\mathcal{P}_x|} \sum_{\bm{x}^+ \in \mathcal{P}_x \in \bm{X}_2\cup\bm{X}_3\cup\bm{X}_4} \exp(\bm{x} \cdot \bm{x}^{+}/\tau)}{\sum\nolimits_{\bm{x}^{-}\in\mathcal{N}_x \in \bm{X}_2\cup\bm{X}_3\cup\bm{X}_4} \exp(\bm{x} \cdot \bm{x}^{-}/\tau)} \tag{2}
\label{cross-scale}
\end{align} 

Therefore, dense connections among different scales enable effective information exchange, which is beneficial for extracting robust forensic features against scale variations.

\noindent 
\textbf{Cross-modality loss.} Inspired by self-supervised learning in the field of natural language processing \cite{gao2021simcse}, cross-modality contrastive learning is further integrated to enhance the forensic discrimination ability. Specifically, the image $X$ is input to the backbone network twice sequentially, resulting in two versions of feature maps, $\bm{X}_i$ and $\hat{\bm{X}}_i$. Due to the randomness of \textit{dropout}, such maps are similar but not identical. Such double encoding expands the number of trainable samples and yields two modalities of training samples without network complexity increase. For a pixel embedding $\bm{x} \in \bm{X}_1$, the positive and negative pixel embedding collections $\mathcal{P}_x$, $\mathcal{N}_x$ are sampled from $\hat{\bm{X}}_1$. Then the cross-modality contrastive loss is adopted as

\begin{align}
\mathcal{L}^{\text{3}}_x= -_{\!}\log\frac{\frac{1}{|\mathcal{P}_x|} \sum_{\bm{x}^+ \in \mathcal{P}_x \in \hat{\bm{X}}_1} \exp(\bm{x} \cdot \bm{x}^{+}/\tau)}{\sum\nolimits_{\bm{x}^{-}\in\mathcal{N}_x \in \hat{\bm{X}}_1} \exp(\bm{x} \cdot \bm{x}^{-}/\tau)} \tag{3}
\label{cross-modality}
\end{align}

Considering computational complexity, we utilize the feature $\bm{X}_1$ with the highest resolution to compute $\mathcal{L}^{\text{1}}_x$ and $\mathcal{L}^{\text{3}}_x$. Overall, the total multi-view contrastive loss $\mathcal{L}^{\text{Contra}}_x$ utilized in training the backbone network is defined as  

\begin{align}
\mathcal{L}^{\text{Contra}}_x = \mathcal{L}^{\text{1}}_x + \mathcal{L}^{\text{2}}_x + \mathcal{L}^{\text{3}}_x \tag{4}
\label{total_loss}
\end{align}

\subsection{Supervised Pixel-wise Forgery Detection}
After the completion of training, the weights of the backbone network are frozen. Then a localization head is fine-tuned project $\{\bm{X}_1, \bm{X}_2, \bm{X}_3, \bm{X}_4\}$ into a score map $\bm{Y}$. Let $y$ be the score vector (activated by \textit{sigmoid}) for pixel $x$, \textit{i.e.}, $y \in \bm{Y}$. Given $y$ for pixel $x$ w.r.t its groundtruth label $\hat{y}\in \{0, 1\}$, the improved cross-entropy loss \cite{lin2017focal} is optimized with 

\begin{align}
    \begin{split}
        \mathcal{L}^{\mathrm{CE}}_x(\hat{y},y)=& \begin{aligned}-\alpha\left(1-y\right)^{\gamma} \times \hat{y}\log\left(y\right)\end{aligned}  \\
        &\begin{aligned}-(1-\alpha)y^{\gamma} \times (1-\hat{y})\log{(1-y)}\end{aligned}
    \end{split} \tag{5}
\label{CE}
\end{align}
\noindent
where $\alpha$ and $\gamma$ are the hyper-parameters, and they are empirically set as 0.5 and 2, respectively. While contrastive learning encourages pixels within an image to cluster according to their labels, CE loss-based fine-tuning rearranges these clusters so that they fall on the correct side of the decision boundary. 

\section{Experiments}
\subsection{Experimental Setup}
\noindent
\textbf{Datasets.} For comprehensive and fair comparisons with existing algorithms, three sets of experiments are conducted on different scales of training datasets, as illustrated in Table \ref{dataset}. The configurations for each experiment are as follows:

\begin{itemize}[leftmargin=*]
    \setlength{\itemindent}{0em}
    \item \textbf{Experiment 1.} \label{exp1} Following prior works \cite{dong2022mvss, kong2023pixel}, our model is trained on the CASIAv2 dataset \cite{dong2013casia}, which comprises 5123 tampered images. Then we test our model on additional 10 datasets, including CASIAv1 \cite{dong2013casia}, NIST \cite{guan2019mfc}, Columbia \cite{hsu2006columbia}, IFC \cite{IFC}, IMD \cite{novozamsky2020imd2020}, Coverage \cite{wen2016coverage}, DSO \cite{carvalho2015illuminant}, DEF-test \cite{mahfoudi2019defacto}, Wild \cite{huh2018fighting}, and Korus \cite{korus2016evaluation}. In \textbf{Experiment 1}, all models are trained exclusively on the CASIAv2 dataset. 
    
    \item \textbf{Experiment 2.} \label{exp2} Following \cite{liu2022pscc, guo2023hierarchical, zeng2024mgqformer, zhang2024catmullrom, zhu2024learning, li2023edge}, we also pre-train our model on a randomly selected set of 60,000 tampered images from the CAT-Net dataset \cite{kwon2022learning}. For subsequently fine-tuning the MPC, we adopt the same training-test ratio configuration as \cite{liu2022pscc, guo2023hierarchical} on NIST, Coverage and CASIA. Lastly, we conduct testing on 4 datasets (\textit{i.e.}, CASIAv1 \cite{dong2013casia}, NIST \cite{guan2019mfc}, IMD \cite{novozamsky2020imd2020} and Coverage \cite{wen2016coverage}). In \textbf{Experiment 2}, all models are trained on the equivalently scaled dataset for fair comparisons. 
    
    \item \textbf{Experiment 3.} \label{exp3} Our model is trained using the same datasets with CAT-Net \cite{kwon2022learning} and TruFor\cite{ guillaro2023trufor}, and tested on other 10 public datasets, including CASIAv1 \cite{dong2013casia}, NIST \cite{guan2019mfc}, Columbia \cite{hsu2006columbia}, Coverage \cite{wen2016coverage}, DSO \cite{carvalho2015illuminant}, Wild \cite{huh2018fighting}, Korus \cite{korus2016evaluation}, MISD \cite{kadam2021multiple}, CoCoGlide \cite{guillaro2023trufor} and FF++ \cite{rossler2019faceforensics++}. Due to limitations in computational resources, we only compare our method with CAT-Net and TruFor, which are trained on the same dataset.
\end{itemize} 

\noindent
\textbf{Metrics.} Similar to previous works \cite{kong2023pixel, wu2022robust, guillaro2023trufor}, the accuracy of pixel-level forgery localization is measured by F1 and IoU. The fixed threshold 0.5 is adopted to binarize the localization probability map. The average F1 and IoU values of each test dataset are reported as the statistical performance of forgery localization algorithms. Considering the obvious imbalance of image sample numbers across different datasets, the average F1 and IoU values of all test datasets are computed as 

\begin{equation}
\mathrm{\mathbf{Average} } = \frac{\sum_{i=1}^{i=N}  \mathrm{Metric}_{D_i}\times \mathrm{Num}_{D_i}}{ {\sum_{i=1}^{i=N}\mathrm{Num}_{D_i} } }  \tag{6}
\label{average}
\end{equation}

\noindent
where $\mathrm{Metric}_{D_i}$ refers to the average metric (F1 or IoU) of the \textit{i}-th dataset. And $\mathrm{Num}_{D_i}$ denotes the number of images contained in the \textit{i}-th dataset. Such a metric is inherently the image sample-level average value, instead of the simple dataset-level arithmetic mean. The computation method in Eq. \ref{average} is suitable for recomputing the sample-level average in existing literatures, in which only the metrics of each dataset $\mathrm{Metric}_{D_i}$ are available.

\noindent
\textbf{Implementation details.} The proposed MPC is implemented by PyTorch.  We train the network on a single A800 GPU for two stages.  In the first training stage, the learning rate starts from 1e-4 and decreases by the Reduce LR On Plateau strategy. And in the second training stage, the learning rate starts from 1e-4 and decreases by the Cosine Annealing strategy.  Adam is adopted as the optimizer, batch size is 4 and all the images used in training are resized to $512 \times 512$ pixels. As \cite{wu2022robust}, the common data augmentations, including flipping, blurring, compression, noising and resizing are adopted.

\subsection{Comparison to State-of-the-Arts}
\noindent
\textbf{Experiment 1.} We compare the performance of our MPC with 16 existing image forgery localization or semantic segmentation algorithms, all of which are trained on the CASIAv2 dataset.

\begin{table*}
  \centering
  \caption{Image forgery localization performance F1[\%] in \hyperref[dataset]{Experiment 1}. The best results are highlighted in \textcolor{red}{red}.}
  \tabcolsep=6 pt
  \begin{adjustbox}{width=\textwidth}
  \begin{tabular}{r|l|cccccccccc|cc}
    \toprule[1.5pt]
    Method & Year-Venue & NIST & Columbia & IFC & CASIAv1 & IMD & Coverage & DSO & DEF-test   &Wild & Korus  & \textbf{Average} \\
    \midrule
    FCN \cite{long2015fully} & 2015-CVPR                  &16.7	&22.3	&7.9	&44.1	&21.0	&19.9	&6.8	&13.0	&19.2	&12.2  & 17.4\\
    
    U-Net \cite{ronneberger2015u} & 2015-MICCAI           &17.3	&15.2	&7.0	&24.9	&14.8	&10.7	&12.4	&4.5	&17.5	&11.7  & 9.6\\
   
    DeepLabv3 \cite{chen2017deeplab} & 2018-TPAMI         &23.7	&44.2	&8.1	&42.9	&21.6	&15.1	&16.4	&6.8	&22.0	&12.0 &14.7\\ 
   
    MFCN \cite{salloum2018image} & 2018-JVCIR             &24.3	&18.4	&9.8	&34.6	&17.0	&14.8	&15.0	&6.7	&16.1	&11.8 &12.7\\ 

    RRU-Net \cite{bi2019rru} & 2019-CVPRW                 &20.0	&26.4	&5.2	&29.1	&15.9	&7.8	&8.4	&3.3	&17.8	&9.7 &9.7\\ 
   
    ManTra-Net \cite{wu2019mantra} & 2019-CVPR             &15.8	&45.2	&11.7	&18.7	&16.4	&23.6	&25.5	&6.7	&31.4	&11.0 &11.7\\ 
   
    HPFCN \cite{li2019localization} & 2019-ICCV           &17.2	&11.5	&6.5	&17.3	&11.1	&10.4	&8.2	&3.8	&12.5	&9.7 &7.6\\ 
    
    H-LSTM \cite{bappy2019hybrid} & 2019-TIP              &\color{red}35.7	&14.9	&7.4	&15.6	&20.2	&16.3	&14.2	&5.9	&17.3	&14.3&11.7\\ 
    
    SPAN \cite{hu2020span} & 2020-ECCV                    &21.1	&50.3	&5.6	&14.3	&14.5	&14.4	&8.2	&3.6	&19.6	&8.6 &8.8\\
    
    ViT-B \cite{dosovitskiy2020vit}    & 2020-ICLR        &25.4	&21.7	&7.1	&28.2	&15.4	&14.2	&16.9	&6.2	&20.8	&17.6 &11.8\\ 
    
    Swin-ViT \cite{liu2021swin} & 2021-ICCV               &22.0	&36.5	&10.2	&39.0	&30.0	&16.8	&18.3	&15.7	&26.5	&13.4 &21.0 \\
    
    PSCC-Net \cite{liu2022pscc} & 2022-TCSVT              &17.3	&50.3	&11.4	&33.5	&19.7	&22.0	&29.5	&7.2	&30.3	&11.4 &14.0\\
    
    MVSS-Net++ \cite{dong2022mvss} & 2022-TPAMI           &30.4	&66.0	&8.0	&51.3	&27.0	&\color{red}48.2	&27.1	&9.5	&29.5	&10.2 & 19.2 \\	
    	
    CAT-Net \cite{kwon2022learning} & 2022-IJCV           &10.2	&20.6	&9.9	&23.7	&25.7	&21.0	&17.5	&20.6	&21.7	&8.5 & 20.6\\
   
    EVP \cite{liu2023evp} & 2023-CVPR                     &21.0	&27.7	&8.1	&48.3	&23.3	&11.4	&6.0	&9.0	&23.1	&11.3  & 16.2 \\
   
    PIM \cite{kong2023pixel} & 2023-arxiv                                      &28.0	&\color{red}68.0	&15.5	&\color{red}56.6	&41.9	&25.1	&25.3	&16.7	&41.8	&23.4 & 26.9\\
    \midrule
    MPC (Ours) & 2024                                           &29.1	&67.6	&\color{red}17.5	&44.8	&\color{red}48.5	&41.0	&\color{red}36.9	&\color{red}22.0	&\color{red}43.3	&\color{red}25.1  & \color{red}30.6\\
    \bottomrule[1.5pt]
  \end{tabular}
  \end{adjustbox}
  \label{f1_table}
\end{table*}

\begin{table*}
  \centering
 \caption{Image forgery localization performance IoU[\%] in \hyperref[dataset]{Experiment 1}. The best results are highlighted in \textcolor{red}{red}.}
 \tabcolsep=6 pt
  \begin{adjustbox}{width=\textwidth}
  \begin{tabular}{r|l|cccccccccc|cc}
    \toprule[1.5pt]
    Method & Year-Venue & NIST & Columbia & IFC & CASIAv1 & IMD & Coverage & DSO & DEF-test   &Wild & Korus  & \textbf{Average} \\
    \midrule
    FCN \cite{long2015fully} & 2015-CVPR           &11.4	&17.7	&5.8	&36.7	&15.8	&11.7	&4.3	&8.9	&14	&8.9	&12.8\\ 		
    
    U-Net \cite{ronneberger2015u} & 2015-MICCAI    &12.8	&9.7	&4.8	&20.4	&10.5	&7.2	&8.2	&3.1	&12.1	&8.2	&7.0\\ 		
    
    DeepLabv3 \cite{chen2017deeplab} & 2018-TPAMI  &19.1	&35.3	&5.8	&36.1	&15.9	&10.6	&11.2	&5.0	&16.2	&8.4	&11.3\\ 			
   
    MFCN \cite{salloum2018image} & 2018-JVCIR     &19.3	&12.3	&7.4	&29.1	&12.4	&10.0	&10.3	&5.0	&11.2	&8.3	&9.7\\ 
    	
    RRU-Net \cite{bi2019rru} & 2019-CVPRW          &15.6	&19.6	&3.9	&24.4	&11.9	&5.7	&5.7	&2.4	&13.1	&6.8	&7.4\\ 
    										
    ManTra-Net \cite{wu2019mantra} & 2019-CVPR    &9.8	&30.1	&6.8	&11.1	&9.8	&13.9	&15.3	&3.9	&20.1	&6.1	&7.0\\  
    
    HPFCN \cite{li2019localization} & 2019-ICCV    &12.6	&7.6	&4.5	&13.7	&7.6	&7.0	&5.4	&2.6	&8.4	&6.4	&5.4\\	
    	
    H-LSTM \cite{bappy2019hybrid} & 2019-TIP       &\color{red}27.6	&9.0	&4.7	&10.1	&13.1	&10.8	&8.4	&3.7	&10.6	&9.4	&7.7\\ 	
    
    SPAN \cite{hu2020span} & 2020-ECCV            &15.6	&39	&3.7	&11.2	&10.0	&10.5	&4.9	&2.4	&13.2	&5.5	&6.2\\ 	
  
    ViT-B \cite{dosovitskiy2020vit} & 2021-ICLR    &19.7	&16.4	&5.1	&23.2	&19.2	&10.1	&12.1	&4.5	&15.2	&13.0	&10.4\\
   	
    Swin-ViT \cite{liu2021swin} & 2021-ICCV        &16.7	&29.7	&7.8	&35.6	&24.3	&12.4	&13.2	&12.9	&21.4	&10.3	&17.3\\
    
    PSCC-Net \cite{liu2022pscc} & 2022-TCSVT       &10.8	&36.0	&6.7	&23.2	&12.0	&13.0	&18.5	&4.2	&19.3	&6.6	&8.8\\  								
   									
    MVSS-Net++ \cite{dong2022mvss} & 2022-TPAMI    &23.9	&57.3	&5.5	&39.7	&20	&\color{red}38.4	&18.8	&7.6	&21.9	&7.5	&14.8\\  
    
    CAT-Net \cite{kwon2022learning} & 2022-IJCV   &6.2	&14.0	&6.2	&16.5	&18.3	&14.1	&11.0	&15.2	&14.4	&4.9	&14.7\\		
     											
    EVP \cite{liu2023evp} & 2023-CVPR              &16.0	&21.3	&6.2	&42.1	&18.3	&8.3	&4.3	&7.0	&18.2	&8.4	&13.0\\				
    
    PIM \cite{kong2023pixel} & 2023-arxiv                               &22.5	&60.4	&11.9	&\color{red}51.2	&34.0	&18.8	&19.4	&13.3	&33.8	&18.2	&22.2\\
    \midrule
    MPC (Ours) & 2024                                    &23.1	&\color{red}61.7	&\color{red}13.3	&41.2	&\color{red}40.1	&30.2	&\color{red}28.8	&\color{red}17.5	&\color{red}34.9	&\color{red}19.1	&\color{red}25.1\\
    \bottomrule[1.5pt]
\end{tabular}
 \end{adjustbox}
  \label{iou_table}
\end{table*}

Tables \ref{f1_table} and \ref{iou_table} show the evaluation results for the pixel-level localization performance of different methods. Overall, our method achieves the highest average localization accuracy, surpassing the second-best PIM \cite{kong2023pixel} by 3.7\%, 2.9\% in terms of F1, IoU, respectively. Additionally, our method achieves the best F1, IoU scores on 6, 7 datasets, respectively. Despite the 10 testing datasets exhibit diverse distributions, the average localization performance of our MPC outperform all previous approaches. Such impressive results demonstrate the high localization precision and the strong generalization ability of our method across diverse forged images. Such advantages are particularly noteworthy for advancing forensic localizers into real-world applications. 

Moreover, we find that when trained only on the small-scale CASIAv2 dataset, localization performance of the simple semantic segmentation network built using Swin-ViT \cite{liu2021swin} surpasses that of most specifically designed forensic algorithms \cite{liu2022pscc, dong2022mvss, hu2020span, liu2023evp, kwon2022learning, wu2019mantra}. Such results indicate that most prior forensic algorithms may not effectively learn the tampering traces they rely on, when the training dataset is not sufficiently large. In contrast, our MPC does not rely on specific forensic clues. Instead, it constructs a structured feature space through multi-view contrastive learning to distinguish between pristine and tampered pixels.

\noindent
\textbf{Experiment 2.} As enforced in Refs. \cite{zhou2018learning, zhou2020generate, hu2020span, hao2021transforensics, liu2022pscc, dong2022mvss, zhuo2022self, wang2022objectformer, li2022image, zeng2023towards, liu2023tbformer, lin2023image, li2023edge, guo2023hierarchical, xu2023up, shi2023transformer, zhang2024catmullrom, zhu2024learning, zeng2024mgqformer}, this experiment typically commences with pre-training on a medium-scale dataset, with 60\textit{k} $\sim$ 100\textit{k} samples, followed by fine-tuning on a subset of the test datasets. 
Due to 
the absence of additional test datasets, such experiments can not reflect the model's generalization performance. Nevertheless, for a fair comparison, we adopt the same experimental setups as the existing 19 image forgery localization methods. 

\begin{table*}
\caption{Image forgery localization performance F1[\%] and IoU[\%] in \hyperref[dataset]{Experiment 3}. The best results are highlighted in \textcolor{red}{red}.}
\begin{adjustbox}{width=\textwidth}
\begin{tabular}{r|l|c|cccccccccccccccccccc|cc}
\toprule[1.5pt]
\multirow{2}{*}{Method} &\multirow{2}{*}{Year-Venue}      &\multirow{2}{*}{\#Params.} & \multicolumn{2}{c}{Columbia} & \multicolumn{2}{c}{DSO} & \multicolumn{2}{c}{CASIAv1} & \multicolumn{2}{c}{NIST} & \multicolumn{2}{c}{Coverage}& \multicolumn{2}{c}{Korus}& \multicolumn{2}{c}{Wild}& \multicolumn{2}{c}{MISD} & \multicolumn{2}{c}{CoCoGlide} & \multicolumn{2}{c|}{FF++} & \multicolumn{2}{c}{\textbf{Average}} \\ \cmidrule{4-25}
                        &   &   & F1  & IoU       & F1    & IoU      & F1   & IoU         & F1   & IoU       & F1   & IoU  & F1 & IoU     & F1    & IoU     & F1     & IoU      & F1     & IoU   & F1     & IoU  & F1     & IoU\\         
\midrule                    
CAT-Net \cite{kwon2022learning}   & 2022-IJCV      &114.3M   &79.3 &74.6 &47.9 &40.9 &71.0 &63.7 &30.2 &23.5 &28.9 &23.0  & 6.1 &4.2 &34.1 &28.9 &39.4 &31.3 &36.3 &28.8   &12.3  &9.5 &37.6 &32.0       \\
TruFor \cite{guillaro2023trufor}    & 2023-CVPR    &68.7M&   79.8 &74.0 &\color{red}91.0 &\color{red}86.5 &69.6 &63.2 &\color{red}47.2 &\color{red}39.6 &52.3 &45.0  &\color{red}37.7&\color{red}29.9&\color{red}61.2 &\color{red}51.9 &60.0 &47.5 &35.9 &29.1   &\color{red}69.2  &\color{red}56.5&59.8 &51.1   \\
MPC (Ours)                    & 2024        &41.8M    &\color{red}94.5 &\color{red}93.6 &51.2 &38.9 &\color{red}74.5 &\color{red}68.6 &43.7 &36.6 &\color{red}61.7 &\color{red}52.6  & 29.6&22.6&59.4 &50.5 &\color{red}72.6 &\color{red}59.8 &\color{red}42.4 &\color{red}33.4   &\color{red}69.2  &54.6&\color{red}61.2 &\color{red}52.0         \\
\bottomrule[1.5pt]
\end{tabular}
\end{adjustbox}
\label{800k}
\end{table*}

Table \ref{ft_f1_table} shows comparison results of the fine-tuned models in Experiment 2. The cases of unavailable results marked by '-' are attributed to the undisclosed source code or no test results given in the literatures. Our MPC achieves the best average F1 score compared to existing methods and outperforms the second-best TBFormer \cite{liu2023tbformer} by 3.8\%. Note that TBFormer does not test on the IMD dataset, which is crucial for evaluating the generalization ability of forgery localization algorithms. The corresponding improvements of MPC can reach more than 24.5\%, compared to PCL \cite{zeng2023towards} which also uses contrastive learning. Overall, our MPC can adapt to datasets with different distributions and achieve competitive localization performance. 

\noindent
\textbf{Experiment 3.} In this set of experiments, we compare our approach with CAT-Net \cite{kwon2022learning} and TruFor \cite{guillaro2023trufor}. All models are trained using the CAT-Net dataset \cite{kwon2022learning}, which consists of over 800\textit{k} tampered images. Table \ref{800k} reports the F1 and IoU of different methods on 10 test datasets. Our MPC outperforms TruFor on the Columbia, CASIAv1, Coverage, MISD and CoCoGlide datasets, while it falls short of TruFor on the remaining datasets. On average, our method outperforms the current best TruFor by 1.4\% and 0.9\% in terms of F1 and IoU. Overall, our MPC demonstrates forgery localization performance comparable to TruFor and surpasses CAT-Net on all datasets. Furthermore, our approach boasts the advantage of fewer parameters. While CAT-Net (114.3M) and TruFor (68.7M) enjoy dual-branch architectures utilizing HRNet \cite{wang2020deep} and SegFormer \cite{xie2021segformer} backbones respectively, our MPC (41.8M) solely leverages a single-branch HRFormer. Our scheme abstains from the design of additional modules, and achieves comparable or even better localization performance at a lower computational cost.

\begin{table}
  \caption{Image forgery localization performance F1[\%] of fine-tuned models in \hyperref[dataset]{Experiment 2}. ‘-’ denotes that the result is unavailable. The best results are highlighted in \textcolor{red}{red}.}
  \tabcolsep=2 pt
  \begin{adjustbox}{width=\linewidth}
  \begin{tabular}{r|l|c|cccc|c}
    \toprule[1.5pt]
    Method & Year-Venue & \#Data & NIST & Coverage & CASIAv1 & IMD & \textbf{Average} \\
    \midrule
    RGB-N \cite{zhou2018learning} & 2018-CVPR             &42\textit{k}	&72.2	&43.7	&40.8	&-	&52.2 \\

    GSR-Net \cite{zhou2020generate} & 2020-AAAI           &-	&73.6	&48.9	&57.4	&51.3	&56.3 \\

    SPAN \cite{hu2020span} & 2020-ECCV  &96\textit{k}	&58.2	&55.8	&38.2	&-	&46.4		\\

    TransForensics \cite{hao2021transforensics} & 2021-ICCV    &-	&-	&64.8	&47.9	&54.5	&52.8 \\

    PSCC-Net \cite{liu2022pscc} & 2022-TCSVT    &100\textit{k}	&74.2	&72.3	&55.4	&-	&63.2		\\

    MVSS-Net++ \cite{dong2022mvss} & 2022-TPAMI   &60\textit{k}	&85.4	&75.3	&54.6	&-	&66.9	\\	

    SAT \cite{zhuo2022self} & 2022-TIFS          &98\textit{k}	&87.8	&\color{red}84.3	&59.2	&-	&71.0	\\
    	
    ObjectFormer \cite{wang2022objectformer} & 2022-CVPR   &62\textit{k}	&82.4	&75.8	&57.9	&-	&67.8	\\

    MSAES \cite{li2022image}   & 2022-TMM          &70\textit{k}	&85.4	&-	&45.6	&-	&60.7	\\

    PCL \cite{zeng2023towards}   & 2023-TCSVT          &100\textit{k}	&78.0	&62.0	&46.7	&50.3	&54.1	\\
    
    TBFormer \cite{liu2023tbformer}  & 2023-SPL       &150\textit{k}	&83.4	&-	&69.6	&-	&74.8	\\ 

    EMT-Net \cite{lin2023image} & 2023-PR         &110\textit{k}	&82.5	&35.3	&45.9	&-	&58.3	\\

    ERMPC \cite{li2023edge} & 2023-CVPR          &60\textit{k}	&83.6	&77.3	&58.6	&-	&68.7	\\

    HiFi-Net \cite{guo2023hierarchical} & 2023-CVPR     &100\textit{k}	&85.0	&80.1	&61.6	&-	&71.1	\\
    
    UP-Net \cite{xu2023up} & 2023-TCSVT      &65\textit{k}	&\color{red}91.6	&57.7	&61.5	&57.4	&63.8	\\
   
    TANet \cite{shi2023transformer} & 2023-TCSVT       &60\textit{k}	&86.5	&78.2	&61.4	&-	&71.4	\\

    CSR-Net \cite{zhang2024catmullrom} & 2024-AAAI       &60\textit{k}	&83.5	&78.0	&58.2	&-	&68.5	\\

    DNG \cite{zhu2024learning} & 2024-AAAI       &60\textit{k}	&86.8	&81.2	&62.1	&-	&72.1	\\

    MGQFormer \cite{zeng2024mgqformer} & 2024-AAAI       &100\textit{k}	&-	&58.8	&-	&-	&58.8	\\
    \midrule
    MPC (Ours) & 2024              &60\textit{k}	&90.6	&80.1	&\color{red}82.4	&\color{red}69.5	&\color{red}78.6	\\
    \bottomrule[1.5pt]
  \end{tabular}
  \end{adjustbox}
  \label{ft_f1_table}
\end{table}

\begin{table}
\caption{Robustness performance F1[\%] and IoU[\%] against online social networks (OSNs) post-processing, including Facebook (Fb), Wechat (Wc), Weibo (Wb) and Whatsapp (Wa). The best results are highlighted in \textcolor{red}{red}.}
\begin{adjustbox}{width=\linewidth}
\begin{tabular}{r|c|cccccccc|cc}
\toprule[1.5pt]
\multirow{2}{*}{Method} & \multirow{2}{*}{OSNs}     & \multicolumn{2}{c}{CASIAv1} & \multicolumn{2}{c}{Columbia} & \multicolumn{2}{c}{NIST} & \multicolumn{2}{c|}{DSO} & \multicolumn{2}{c}{\textbf{Average}} \\ \cmidrule{3-12}
                        &                           & F1           & IoU          & F1            & IoU          & F1          & IoU         & F1           & IoU     & F1           & IoU     \\         
\midrule                    
CAT-Net \cite{kwon2022learning}                 & \multirow{3}{*}{Fb} & 63.3   & 55.9    & 91.8   &90.0   & 15.1   &11.9   & 12.1   &9.8   &47.4  &42.2    \\
TruFor \cite{guillaro2023trufor}                &                     &67.2    &60.5     & 74.9   &67.1   &35.3    &27.8   & \color{red}65.4   &\color{red}55.2  &57.5  &50.2    \\
MPC (Ours)                                            &                     &\color{red}70.9    &\color{red} 64.4    &\color{red} 95.9   &\color{red}95.2   &\color{red}43.1    & \color{red}35.9  & 51.6   &39.3  &\color{red}63.1  &\color{red}56.6    \\ 
\midrule
CAT-Net \cite{kwon2022learning}                 & \multirow{3}{*}{Wc} & 13.9   & 10.6    & 84.8   &80.8   & 19.1   & 14.9  & 29.9   &26.8  &23.0 &19.4    \\
TruFor \cite{guillaro2023trufor}                &                     & 56.9   & 50.8    & 77.3   &70.3   & 35.1   & 27.4  & 43.6   &31.4  &51.0 &43.9               \\
MPC (Ours)                                            &                     & \color{red}61.3   & \color{red}53.9    &\color{red} 94.7   &\color{red} 93.9  &\color{red} 42.5   & \color{red}35.2  &\color{red} 50.2   &\color{red}37.8  &\color{red}57.6 &\color{red}50.6    \\ 
\midrule
CAT-Net \cite{kwon2022learning}                 & \multirow{3}{*}{Wb} &42.5    &36.2     & 92.1   & 89.7  & 20.8   & 16.0  & 39.4   &35.8  &39.9  &34.6     \\
TruFor \cite{guillaro2023trufor}                &                     &63.7    &57.6     & 80.0   & 73.1  & 33.2   & 26.2  & 46.4   &36.3  &54.3  &47.6      \\
MPC (Ours)                                            &                     &\color{red}70.7    &\color{red}64.6     &\color{red}94.9    &\color{red}94.3   &\color{red} 43.5   & \color{red}36.3  & \color{red}51.3   &\color{red}39.3  &\color{red}63.0  &\color{red}56.7        \\ 
\midrule
CAT-Net \cite{kwon2022learning}                 & \multirow{3}{*}{Wa} & 42.3   & 37.8    & 92.1   &89.9   & 20.1   & 16.8  & 39.2   &36.5  & 39.5 & 35.7             \\
TruFor \cite{guillaro2023trufor}                &                     & 66.3   & 59.9    & 74.7   &66.7   & 32.3   & 25.6  & 37.6   &28.9  & 54.4 & 47.7                 \\
MPC (Ours)                                            &                     & \color{red}70.0   & \color{red}63.4    & \color{red}95.0   & \color{red}94.3  &\color{red}44.0    & \color{red}36.5  & \color{red}52.6   &\color{red}40.4  & \color{red}62.9 &\color{red} 56.2        \\
\bottomrule[1.5pt]
\end{tabular}
\end{adjustbox}
\label{osn}
\end{table}

\subsection{Robustness Evaluation}
We first assess the robustness of image forgery localization methods against the complex post operations introduced by online social networks (OSNs).  Following the prior work \cite{wu2022robust}, the four forgery datasets transmitted through Facebook, Weibo, Wechat and Whatsapp platforms are tested. Table \ref{osn} shows that, our method mostly achieves the highest accuracy across the four datasets for each social network platform. Among the compared methods, MPC enjoys the smallest performance loss incurred by OSNs. Note that CAT-Net achieves higher performance on the processed Columbia dataset due to its specialized learning of JPEG compression artifacts. However, it shows obvious performance decline on other datasets, for instance, with F1=91.8\% on the Facebook version of Columbia while with F1=13.9\% on the Wechat version of CASIAv1. In contrast, our MPC consistently keeps high localization accuracy across all OSN transmissions. Such results verify the robustness of MPC against the online social networks processing.

\begin{figure}[!t]
\centering
\includegraphics[width=\linewidth]{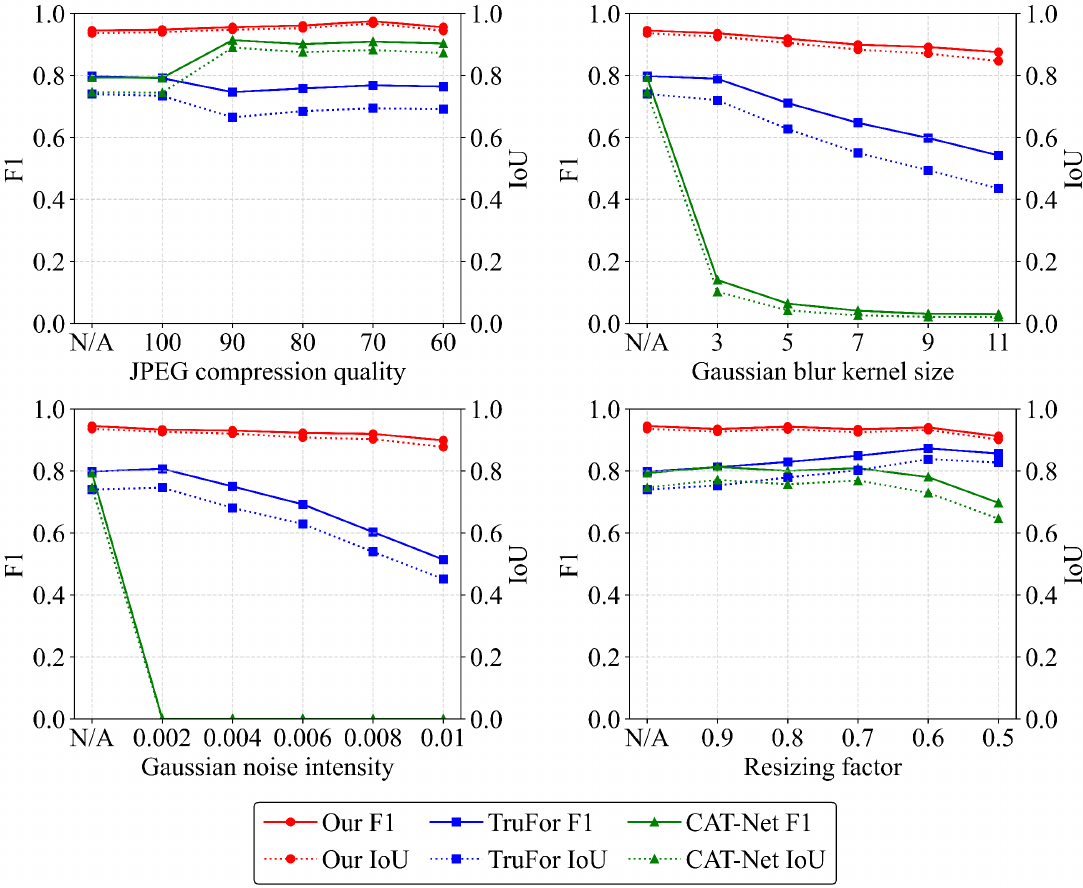}
\caption{Robustness against different post-processing manipulations on the Columbia dataset.}
\label{Post}
\end{figure}

Following the prior works \cite{dong2022mvss, liu2022pscc}, the robustness against post JPEG compression, Gaussian blur, Gaussian noise and resizing is also evaluated on the Columbia dataset.  The results shown in Fig. \ref{Post} verify the high robustness of our MPC against such post-processing. MPC performs the best in all cases. Specially, the blur and noise manipulations greatly reduce the performance of CAT-Net and TruFor, but have minimal impact on our method. Take the noise for example, although F1 values of CAT-Net and TruFor decrease from 0.8 to 0 and 0.5, respectively, that of MPC always keeps above 0.9 across different noise intensities. 

\begin{figure}[!t]
\centering
\includegraphics[width=\linewidth]{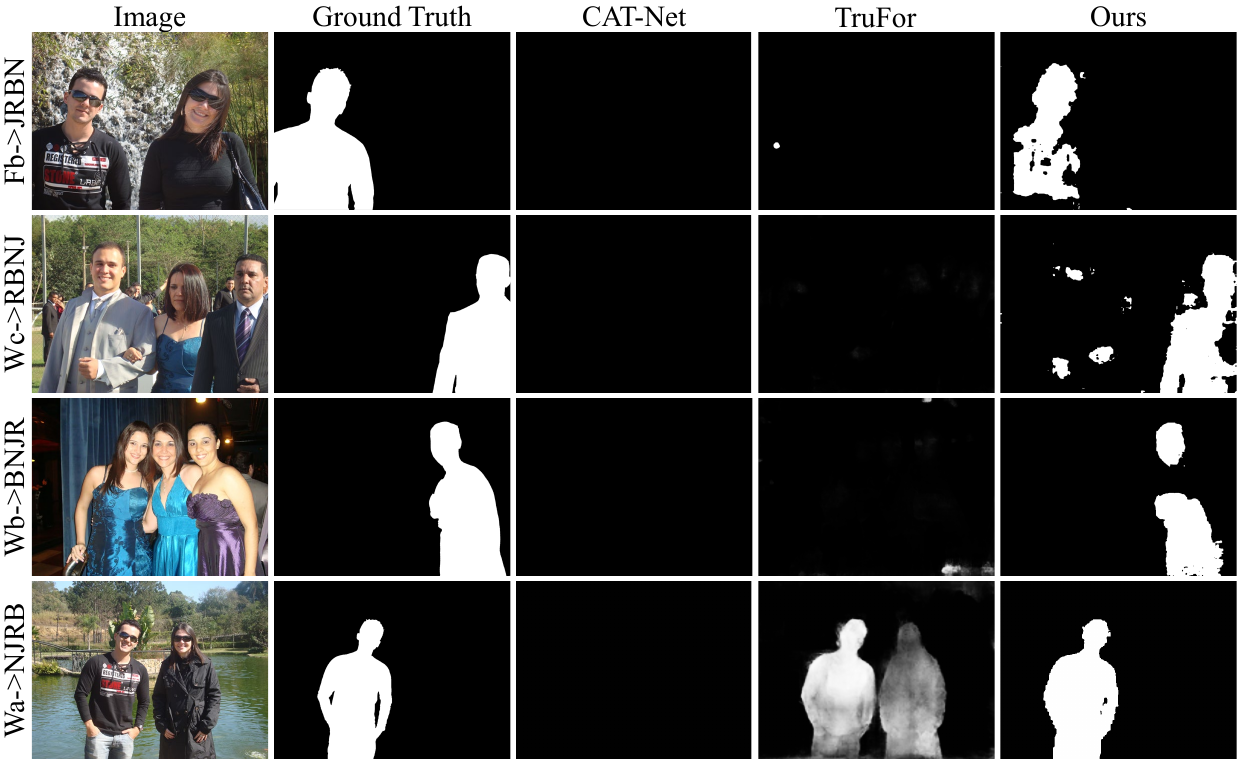}
\caption{Visualization of robustness against combined post-processing manipulations on the DSO dataset.}
\label{vis_OSN_Post}
\end{figure}

We further validate the robustness of our MPC against combined post-processing manipulations. Specifically, We apply additional JPEG compression, Gaussian blur,Gaussian noise and resizing to the Columbia dataset after the OSN  transmission. Furthermore, the order of manipulations is random for simulating real-world post-processing chains. Table \ref{mix} reports the corresponding statistical test result, and Figure \ref{vis_OSN_Post} illustrates qualitative evaluation results on example test images. It can be observed that CAT-Net fails to withstand any combination of post-processing. TruFor's localization performance is also severely compromised, rendering it practically unusable. In contrast, our MPC exhibits considerably strong robustness, surpassing CAT-Net and TruFor by a large margin. Such results are attributed to CAT-Net and TruFor's reliance on specific tampering traces, such as JPEG artifacts and noise inconsistency. Our approach is dedicated in enhancing the intra-class compactness and inter-class separability between pristine and tampered pixels via contrastive learning. Additionally, the data augmentation strategy adopted during training can also enhanced the robustness of our MPC.

\begin{table}
\caption{Robustness against complex combined post-processing manipulations. J, R, B, N represent JPEG compression (quality=60), Resizing (factor=0.6), Gaussian blur (kernel size=5), and Gaussian noise (intensity=0.006), respectively. The best results are highlighted in \textcolor{red}{red}.}
\begin{adjustbox}{width=\linewidth}
\begin{tabular}{r|c|cccccccc}
\toprule[1.5pt]
\multirow{2}{*}{Method} & \multirow{2}{*}{dataset}     & \multicolumn{2}{c}{Fb->JRBN} & \multicolumn{2}{c}{Wc->RBNJ} & \multicolumn{2}{c}{Wb->BNJR} & \multicolumn{2}{c}{Wa->NJRB} \\ \cmidrule{3-10}
                        &                           & F1           & IoU          & F1            & IoU          & F1          & IoU         & F1           & IoU         \\         
\midrule                    
CAT-Net \cite{kwon2022learning}                 & \multirow{3}{*}{Columbia} & 0.4   & 0.2    & 2.2   &1.5   & 0.1   &0.1   & 0.8   &0.5       \\
TruFor \cite{guillaro2023trufor}                &                     &1.8    &1.1     & 7.9   &5.5   &14.7    &12.1   & 51.1   &44.1      \\
MPC (Ours)              &                     &\color{red}49.4    &\color{red} 42.6    &\color{red} 52.2   &\color{red}44.6   &\color{red}80.1    & \color{red}76.0  & \color{red}79.9   &\color{red}77.0  \\ 
\midrule
CAT-Net \cite{kwon2022learning}                 & \multirow{3}{*}{DSO} &0.1    &0.1    &0    &0  &0    &0   &0.6    &0.4      \\
TruFor \cite{guillaro2023trufor}                &                     &2.6    &1.5    &3.1    &1.8  &6.0    &3.8   &36.1    &25.0       \\
MPC (Ours)                        &        & \color{red}36.6   & \color{red}25.6    &\color{red} 32.8   &\color{red}22.0  &\color{red} 44.9   & \color{red}33.9  &\color{red}49.4   &\color{red}38.5    \\ 
\bottomrule[1.5pt]
\end{tabular}
\end{adjustbox}
\label{mix}
\end{table}

\begin{table}
\caption{Comparison of localization performance for ablation studies. MPC is trained on the CASIAv2 dataset. Metric values are in percentage. The best results are highlighted in \textcolor{red}{red}.}
\begin{adjustbox}{width=\linewidth}
\begin{tabular}{c|c|cccc|cc}
\toprule[1.5pt]
\multirow{2}{*}{}                  & \multirow{2}{*}{variant} & \multicolumn{2}{c}{Columbia}  & \multicolumn{2}{c|}{Coverage} & \multicolumn{2}{c}{\textbf{Average}} \\ \cmidrule{3-8}
                                   &                          & F1           & IoU                     & F1           & IoU   & F1           & IoU        \\
\midrule
\multirow{3}{*}{Loss}              & w/o $\mathcal{L}^{\text{1}}_x$    &14.9   &9.3     &9.8   &5.7   &12.9   &7.9            \\
                                   & w/o $\mathcal{L}^{\text{2}}_x$    &56.4   &48.4    &38.3  &28.5  &49.4   &40.7             \\
                                   & w/o $\mathcal{L}^{\text{3}}_x$    &54.1   &46.5    &35.7  &27.7  &47.0   &39.3            \\
\midrule
\multirow{2}{*}{Backbone} & HRFormer-small           &42.0  &31.7      &29.1   &20.2  &37.0   &27.3             \\
                        
                                   & HRFormer-base (MPC)  &\color{red}67.6  &\color{red}61.7  &\color{red}41.0 &\color{red}30.2 &\color{red}57.4 &\color{red}49.6\\
\bottomrule[1.5pt]
\end{tabular}
\end{adjustbox}
\label{Ablation}
\end{table}

\subsection{Ablation Studies}
We conduct extensive ablation studies to validate the effectiveness of our proposed MPC. A summary of the involved sub-experiments is shown in Table \ref{Ablation}.  

\noindent
\textbf{Effectiveness of multi-view pixel contrast.} We remove one of the three contrastive losses and verify the performance improvement from each by gauging the decrease in localization performance. In MPC, within-image loss $\mathcal{L}^{\text{1}}_x$ is the most critical;  its absence causes a catastrophic collapse in localization performance. This is because improving inter-class compactness and intra-class separability through within-image contrast losses is fundamental to our work. Additionally, the cross-modality loss $\mathcal{L}^{\text{3}}_x$ yields more performance benefits than the cross-scale loss $\mathcal{L}^{\text{2}}_x$. Such a result is attributed to the employed \textit{dropout} mechanism, which introduces challenging instances and therefore enhances the model's ability to discern tampered pixel regions.

\noindent
\textbf{Influence of backbone network.} We also investigate the impact of network size on performance. Despite the increased network complexity with HRFormer-base (41.8M) compared to HRFormer-small (8.2M), notable performance gains are achieved. Such improvement is attributed to the larger models with more parameters may enjoy stronger representation learning ability. However, compared to CAT-Net (114.3M) and TruFor (68.7M), our MPC (41.8M) remains lightweight.

\begin{figure*}[!t]
\centering
\includegraphics[width=\textwidth]{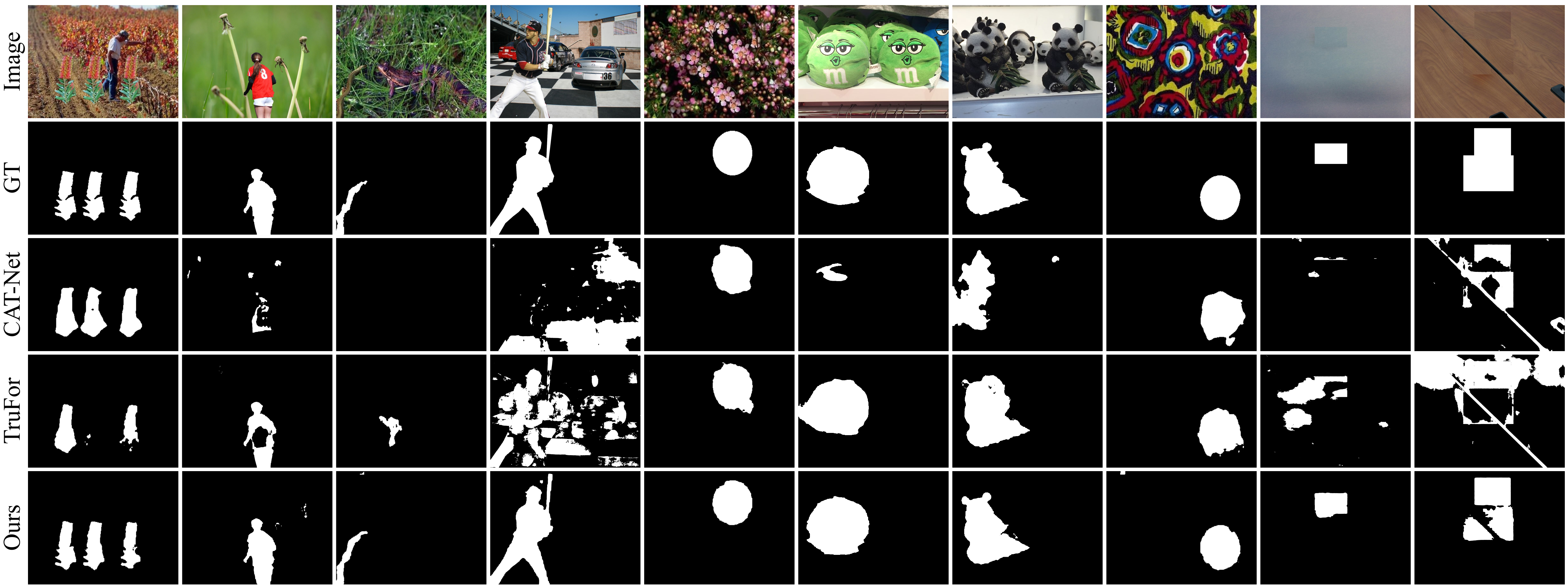}
\caption{Qualitative comparison of forgery localization on some representative testing images. From left to right: four splicing images, four copy-move images, and two removal images. From top to bottom: tampered image, ground truth (GT), and the
localization results from CAT-Net, TruFor and our MPC.}
\label{Common}
\end{figure*}

\begin{figure}[!t]
\centering
\includegraphics[width=\linewidth]{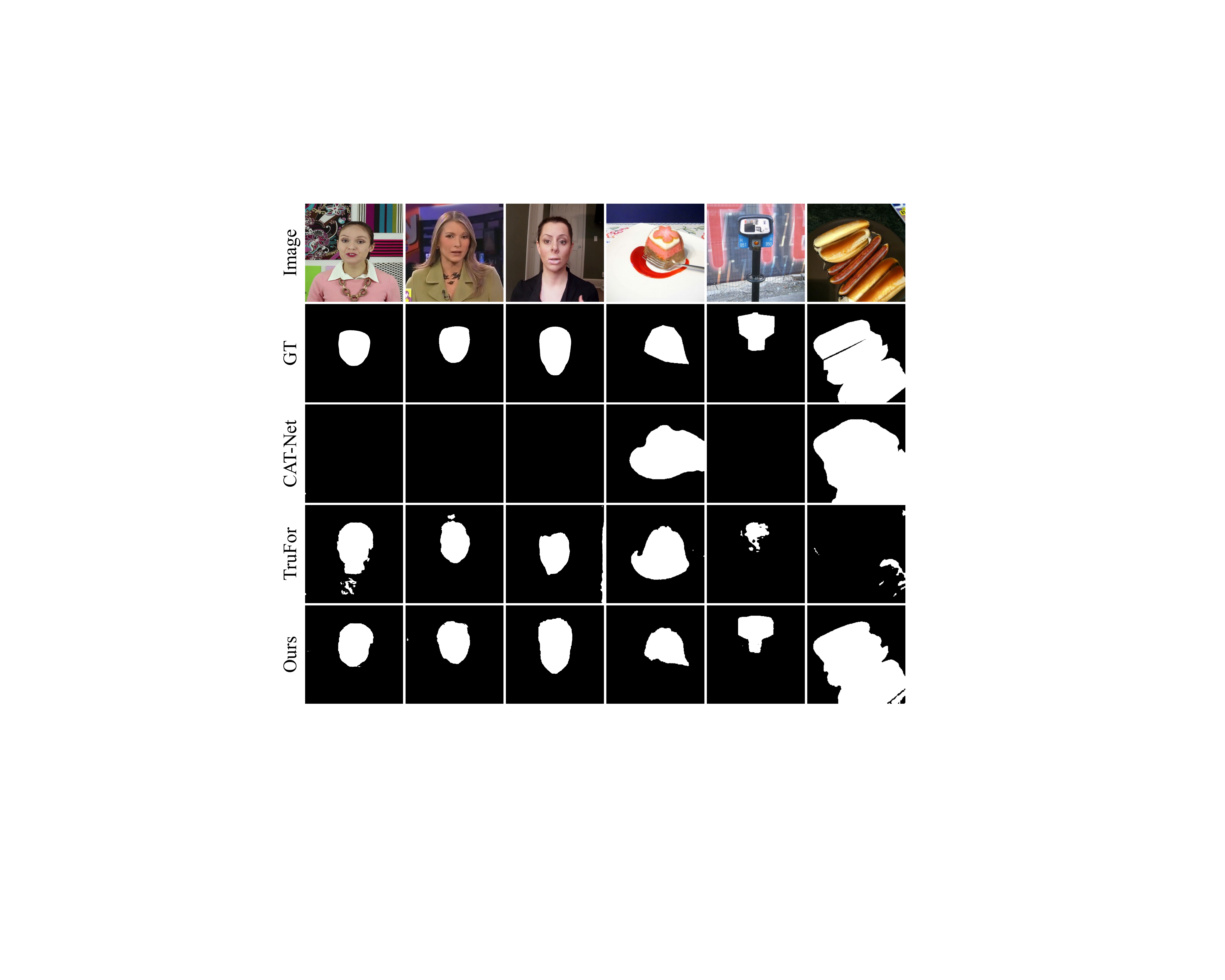}
\caption{Qualitative comparison of forgery localization on some novel tampered images. From left to right: three deepfake images and three local AI-generated images. From top to bottom: tampered image, ground truth (GT), and the
localization results from CAT-Net, TruFor and our MPC.}
\label{AI}
\end{figure}

\subsection{Qualitative Results}
We also qualitatively compare the localization performance of different methods. Several samples are selected from test datasets for encompassing the three popular tampering types, \textit{i.e.}, splicing, copy-move and removal. Fig. \ref{Common} shows the predicted pixel-level forgery localization maps of different methods on the example images. It can be seen that our method produces more accurate localization results for different types of forged images. In most cases, the other localization methods merely detect parts of tampered regions with more or less false alarms. 

In addition to traditional image forgeries, we further qualitatively compare the localization algorithms on some new types of forged images. Specifically, some deepfake and local AI-generated images are selected from the FF++ \cite{rossler2019faceforensics++} and CoCoGlide \cite{guillaro2023trufor} datasets, respectively. The synthetic face images in FF++ are generated using face swapping algorithms such as Face2Face \cite{thies2016face2face}. The locally AI-generated images in CoCoGlide are generated by GLIDE diffusion model \cite{nichol2022glide}. As depicted in Figure \ref{AI}, despite the absence of such type images during training, our method is still able to detect the manipulated regions accurately. In contrast, TruFor tends to produce more false positives and misses, while CAT-Net almost fails completely. Such results demonstrate the strong generalization ability of our MPC, which benefits from the proposed multi-view pixel-wise contrastive learning.

\section{Conclusion}

In this paper, we propose a novel scheme called MPC for trusted image forgery localization. The backbone network is first trained using pixel-wise supervised contrastive loss, which models the pixel relationships in the feature space from three perspectives: within-image, cross-scale, and cross-modality. Then the localization head is fine-tuned via CE loss, resulting in a better pixel localizer. Comprehensive and fair comparisons with existing image forgery localization methods are conducted through three sets of mainstream experiments. Extensive experiments demonstrate that our approach achieves superior generalization ability and robustness compared to the state-of-the-arts. The MPC exhibits strong robustness against the challenging post-processing of online social networks and complicated operation chains. In future work, we aim to investigate more powerful forensic algorithms for addressing extremely challenging forgeries, such as low-light images and new type AI-generated ones.


\bibliographystyle{ACM-Reference-Format}
\bibliography{sample-base}

\end{document}